\documentclass[10pt,twocolumn,letterpaper]{article}

\usepackage{wacv}
\usepackage{times}
\usepackage{epsfig}
\usepackage{graphicx}
\usepackage{amsmath}
\usepackage{amssymb}
\usepackage{stfloats}
\usepackage{subcaption}
\usepackage[accsupp]{axessibility}  
\usepackage[skip=3.5pt]{caption}

\def\wacvPaperID{1328} 

\wacvfinalcopy 

\ifwacvfinal
\fi

\ifwacvfinal
\usepackage[breaklinks=true,bookmarks=false]{hyperref}
\else
\usepackage[pagebackref=true,breaklinks=true,colorlinks,bookmarks=false]{hyperref}
\fi

\begin{document}

\title{Controlled GAN-Based Creature Synthesis via a Challenging Game Art Dataset - Addressing the Noise-Latent Trade-Off
\\[-1.0ex] 
}

\author{Vaibhav Vavilala and David Forsyth\\
University of Illinois at Urbana-Champaign\\
{\tt\small \{vv16, daf\}@illinois.edu}
\\[-1.0ex] 
}


\twocolumn[{%
\renewcommand\twocolumn[1][]{#1}%
\maketitle
\begin{center}
    \centering
    \captionsetup{type=figure}
    \includegraphics[width=.9\textwidth]{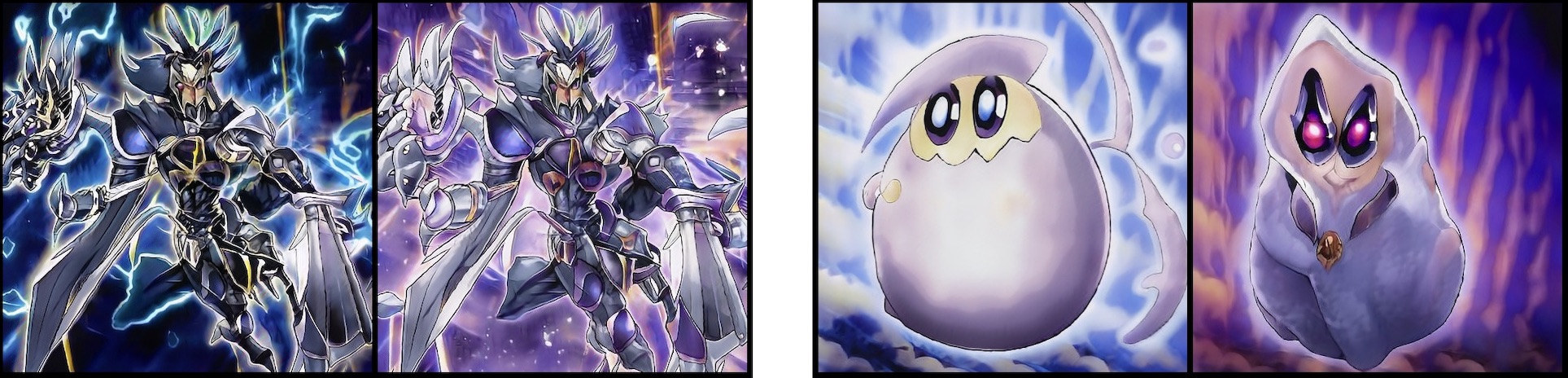}
    \captionof{figure}{StyleGAN representations enable a range of important and useful artist edits in numerous image domains including card art. However, current practices make changing creature identity difficult, often altering card style without materially affecting identity (left pair). Our approach addresses this issue, enabling meaningful edits to card identity (right pair).}

\end{center}%
}]


\begin{figure*}[t!]
\begin{center}

\includegraphics[width=0.95\linewidth]{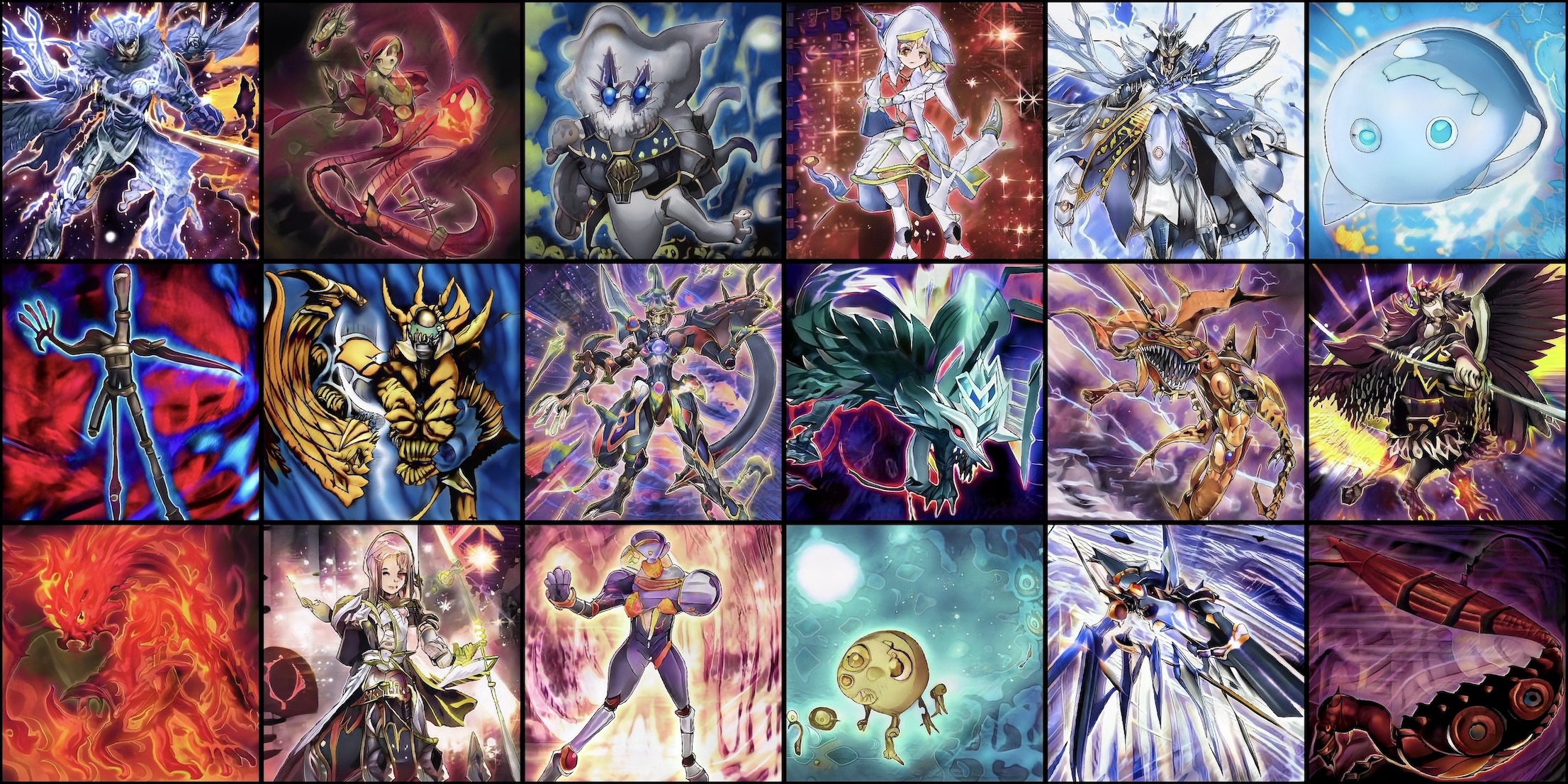}
\end{center}
   \caption{Curated samples generated by a GAN trained on card art. No latent truncation or subsequent editing was applied. AI-assisted art can be used for exploration and even for final quality products. All figures best viewed online, zoomed-in.}
\label{fig:gen}
\end{figure*}
\begin{abstract}

The state-of-the-art StyleGAN2 network supports powerful methods to create and edit art, including generating random images, finding images ``like'' some query, and modifying content or style. Further, recent advancements enable training with small datasets. We apply these methods to synthesize card art, by training on a novel Yu-Gi-Oh dataset. While noise inputs to StyleGAN2 are essential for good synthesis, we find that coarse-scale noise interferes with latent variables on this dataset because both control long-scale image effects. We observe over-aggressive variation in art with changes in noise and weak content control via latent variable edits. Here, we demonstrate that training a modified StyleGAN2, where coarse-scale noise is suppressed, removes these unwanted effects. We obtain a superior FID; changes in noise result in local exploration of style; and identity control is markedly improved. These results and analysis lead towards a GAN-assisted art synthesis tool for digital artists of all skill levels, which can be used in film, games, or any creative industry for artistic ideation.

\end{abstract}


\section{Introduction}
\label{sec:intro}
We show that a change in StyleGAN training procedure results in significant improvements in properties valuable to artists when training on small and sparse datasets. We propose a new dataset, the Yu-Gi-Oh card art dataset, that consists of approx. 11k samples spanning dozens of classes and styles (including humanoid characters, machines, weapons, natural scenes, animals, and mythical beasts). The sheer diversity in identity, pose, lighting, texture, and style makes the dataset appealing but also challenging to model, leading to a natural next step for generative modeling to tackle.

In an artist-based workflow, a key user need is to control content, in this case creature identity. Conventional methods for doing so are ineffective for small datasets as we show in Section~\ref{sec:train}. This is because coarse-scale noise behaves like a latent variable, controlling long-scale features of the image and strongly influencing creature identity as Fig.~\ref{fig:noiseDiff} demonstrates. We show that by suppressing coarse-scale noise variables we obtain significant improvements in quality of synthesis (Section~\ref{sec:betSyn}) and control (\ref{sec:control}).



In the remainder of this paper, we first discuss related approaches using GANs and sparse datasets (Section~\ref{sec:related}). We then describe our data preparation process as well as training details in Section~\ref{sec:data}. In Section~\ref{sec:train}, we explain why our approach is needed to maximize synthesis quality and control, using both quantitative and qualitative metrics. From there, we show in Section~\ref{sec:retain} that key properties inherent in StyleGAN like GAN inversion and style-mixing are retained - these will be useful in any production scenario. We then document how to build an artist workflow to synthesize new art in a controlled fashion in Section~\ref{sec:aw}. Such a workflow can be used by artists to quickly generate new character concept art. Finally, we end with a discussion and conclusion in Sections~\ref{sec:disc} and \ref{sec:conc} respectively.



\section{Related Work}
\label{sec:related}

\subsection{Image Modeling with GANs}
\label{sec:imModel}
Since the introduction of GANs ~\cite{goodfellow2014generative,wang2020state}, generative models have shown to be highly fruitful in synthesizing novel samples from many distributions including faces ~\cite{Karras2020ada,Karras_2019_CVPR}, landscapes ~\cite{radford2015unsupervised}, animals ~\cite{zhang2019self}, and anime ~\cite{jin2017towards}. In many of these domains, the quality of the synthesis has reached a level that is indistinguishable from the training set as measured by FID ~\cite{heusel2017gans} and other image quality metrics. In nearly all these domains, the dataset consists of a single class that is well-posed. For example, the FFHQ dataset includes 70k nicely-cropped human faces. Even in the low-data regime, GANs have have successfully modeled the distribution as long as the dataset is class-consistent and well-posed. However, there is limited work on modeling distributions with diverse poses and identities as we do here.

In a StyleGAN model, a random vector controls the identity of the image; the code is fed into a mapping network consisting of fully connected layers to obtain a latent code in an extended space. A CNN synthesis network consists of a $4\times4$ layer and two layers each from $8\times8$ resolution up to the target resolution. The extended latent code is fed in at each layer to control the synthesis via weight demodulation. StyleGAN also possesses several desirable properties including projection (also known as GAN inversion) whereby the latent code of an existing image can be recovered ~\cite{abdal2019image2stylegan,lipton2017precise}; style-mixing whereby portions of the latent codes from different images can be mixed; and controlled synthesis by perturbing latent codes in important directions like the network weights' eigenvectors to obtain semantically meaningful changes in the output ~\cite{2020ganspace,shen2020closed}.

A follow-up work to StyleGAN dubbed FastGAN ~\cite{liu2021towards} proposes reduced capacity and an architectural modification called ``skip layer excitation modules'' which are variants of skip connections for faster training and smaller data requirements. That work also suggests using discriminator augmentations which have shown to significantly reduce the number of training samples required for GAN convergence from the hundreds of thousands to just a few hundred~\cite{Karras2020ada,zhao2020differentiable}. Our testing of this work did not show any quality improvements over StyleGAN2, but we did observe drastically faster training times as promised.

Transformers ~\cite{vaswani2017attention} have also produced convincing results for image synthesis tasks by predicting tokens in a latent code for a convolutional decoder to use for generation ~\cite{esser2020taming}. Transformer-based synthesis can condition the input on edges, semantic maps, or other class-conditional cues. Our testing of Transformer-based synthesis on the Yu-Gi-Oh dataset did not yield material improvements over StyleGAN2, and key paradigms like style-mixing and PCA editing are not as transparent in this work. 


\begin{figure*}[t]
\begin{center}

\includegraphics[width=0.95\linewidth]{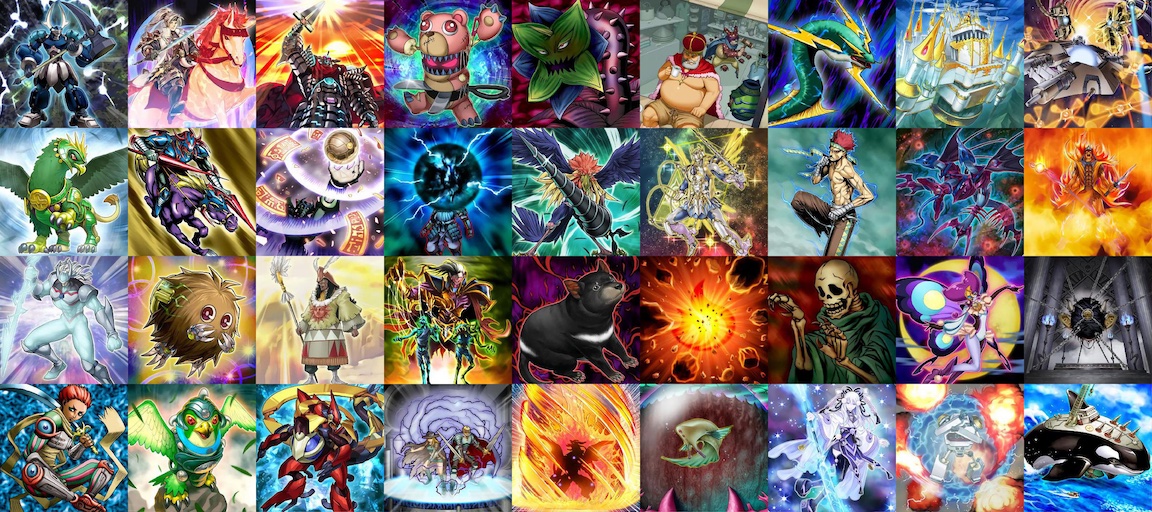}
\end{center}
   \caption{Example images from the Yu-Gi-Oh card art dataset (approx. 11k total samples; 7k monsters-only). Diverse identities, poses, and styles make the dataset appealing but also difficult to model.}
\label{fig:long}
\end{figure*}

\subsection{GAN-Assisted Artist Workflows}
\label{sec:ganAs}
Developing artist-friendly workflows for GANs to assist in image synthesis is a recent and fast-growing area. ThisXDoesNotExist, where X could be faces, dogs, pottery, or a host of other image domains, is a family of websites that randomly generate fake images from a target domain. Artbreeder productionizes a handful of GAN models (anime, human faces, animals, real-world objects) using BigGAN and StyleGAN. Users can randomly generate, interpolate, and project existing images for subsequent editing in an easy to use web app. Artbreeder is closest to the workflow we describe here, the key differences being the choice of dataset, our proposed workflow exposing more controls, and this paper's documentation of how to productionize a GAN. Further, this work identifies and addresses a problem with previous approaches to noise injection without which GAN editing would be hindered with sparse datasets.

\subsection{Sparse Anime Datasets}
\label{sec:sparseData}

Modeling card art is extremely challenging due to the limited data and massive diversity in creature identity and style, as we show in Fig.~\ref{fig:long}. Consequently, the sparsity and absence of a perceptually-aligned classifier means that Instance Selection does not work as shown in Section~\ref{sec:dataCol}. 

On the Yu-Gi-Oh dataset specifically, each card has several metadata associated with it including card type (spell, trap, monster) and description (which is the card effect for all but normal monsters). If the card is a monster, metadata include number of stars, attack \& defense, attribute (light, dark, wind etc.), and monster type (warrior, zombie, dragon, spellcaster etc.). Artistically, monsters generally have a creature overlaid on a colorful background. Spell and trap card art is much more sparse, often including creatures, natural scenes, and weapons. Our highest quality networks use monsters only, as we describe in Section~\ref{sec:trainDet}.

We are aware of one other attempt at using GANs to model Yu-Gi-Oh art from 2018, for which the quality is not strong ~\cite{yu-gan-oh_2018}, and no dataset nor quantitative metric was released. There was also a recent attempt at modeling a similarly sparse dataset of Pokemon (animated creatures) using StyleGAN2 ~\cite{karras2020analyzing} which shows some promise but is nowhere near solved in terms of visual fidelity ~\cite{pokegan}. In particular, capabilities of the GAN like projection, style-mixing, and latent perturbation were not analyzed as we do here. The Danbooru2017 dataset of ~220k well-posed (human) anime faces has been successfully modeled with StyleGAN with very strong quality~\cite{danbooru}. Follow-up work has expanded the dataset to more than 4M images, with the subject matter largely focused on humans. 

\section{Dataset and Training}
\label{sec:data}



\subsection{Data Collection and Processing}
\label{sec:dataCol}
\textit{Collection:} To obtain data, we downloaded ~11k Yu-Gi-Oh cards from the YGOPRO database~\url{https://db.ygoprodeck.com/api-guide/}. We then extracted a $320\times320$ square with just the art, and downsampled to $256\times256$ since powers of two are required for StyleGAN2.

\begin{figure}
\begin{subfigure}{\linewidth}
  \centering
  \includegraphics[width=0.95\linewidth]{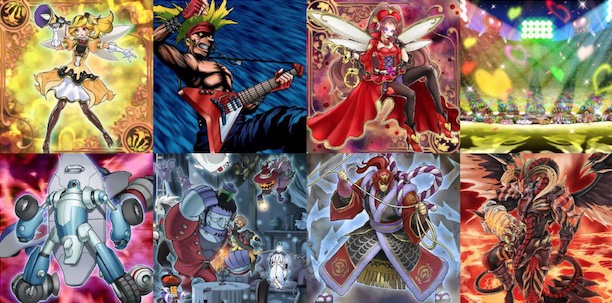}
  \caption{Sparse samples detected by Instance Selection}
  \label{fig:IS}
\end{subfigure}%
\\
\begin{subfigure}{\linewidth}
  \centering
  \includegraphics[width=0.95\linewidth]{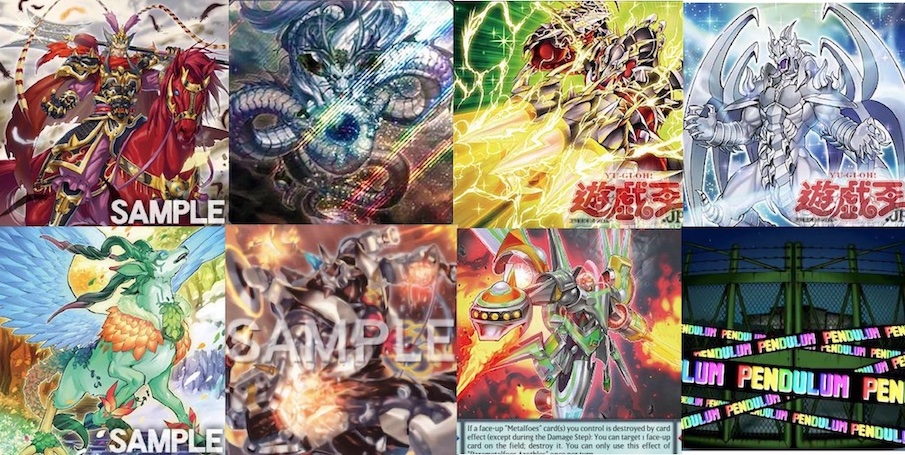}
  \caption{Manually pruned samples}
  \label{fig:manual}
\end{subfigure}
\caption{Instance Selection did not suggest helpful samples to prune (\ref{fig:IS}), so we manually pruned samples (\ref{fig:manual}). We hypothesize that the absence of a perceptually-aligned embedding function accounts for the poor IS results.}
\label{fig:IS_manual}
\end{figure}

\textit{Cleaning:} We resorted to manual cleaning because Instance Selection~\cite{devries2020instance}, a technique that identifies sparse regions of the data manifold, was not successful. Instance Selection applied to card art tends to identify samples, that to a human observer, should not be removed (shown in Fig.~\ref{fig:IS}), and fails to identify samples that should be removed (Fig~\ref{fig:manual}). This is true for multiple embeddings and pretraining configurations (including Inception and ResNet50). We speculate that this is because card art is not perceptually-aligned with the images used to train these classifiers. We thus manually removed approx. 500 samples with unwanted overlaid text or poor scanning (Fig.~\ref{fig:manual}).

\begin{figure}
\begin{subfigure}{0.5\linewidth}
  \centering
  \includegraphics[width=0.95\linewidth]{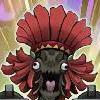}
  \caption{Source jpeg at 320-res\\shows noise/block artifacts}
  \label{fig:jpgNoise}
\end{subfigure}%
\begin{subfigure}{0.5\linewidth}
  \centering
  \includegraphics[width=0.95\linewidth]{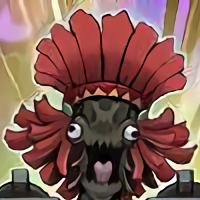}
  \caption{Upscaled + denoised 640-res output}
  \label{fig:upres}
\end{subfigure}
\caption{In addition to training a 256-res network, we applied jpeg denoising/deblocking and super resolution on inputs (\ref{fig:jpgNoise}) to obtain a clean high res output at 512-res (\ref{fig:upres}). Note that the 256/512 res training samples were generated by bicubic downsampling the 320/640 sources.}
\label{fig:postProc}
\end{figure}

\textit{Post-Processing:} Since StyleGAN2 requires image resolutions in powers of 2, we bicubic downsample the images to create a 256-res dataset. We also create a 512 version by running a 2x super resolution network~\cite{ahn2018fast,dong2015image} trained on anime and downsampling from 640 to 512. These networks simultaneously perform jpeg denoising/deblocking as shown in Fig.~\ref{fig:postProc}. While early super resolution work did not generalize well to real-world images due to their emphasis on a single degradation operator (typically bicubic)~\cite{wang2018esrgan,wang2018fully}, recent work (dubbed blind super resolution) employs multiple degradations like Lanczos and bilinear that produce networks with better performance on real-world images~\cite{cornillere2019blind}.

\subsection{Training Details}
\label{sec:trainDet}
For implementation we start with the official NVIDIA StyleGAN2 repository in PyTorch:~\url{https://github.com/NVlabs/stylegan2-ada-pytorch}, which uses mixed precision training. We train a 512-res network on monsters only (7k samples) using four NVIDIA A100 40GB GPUs with a batch size of 96 and 227 hours of training time for 25M images. The generator and discriminator have 28.7M and 28.9M params respectively, and we choose the default hyperparameter for R1 regularization. We found it useful to implement some learning rate decay to help the network converge. Our best run achieved a FID of $10.73$ (Table~\ref{table:fid1}). Our inference time on one A100 GPU is $<0.05$ sec per image, which is suitable for interactive workflows.

\begin{figure*}[h]
\begin{subfigure}[b]{0.5\linewidth}
  \centering
  \includegraphics[width=0.9\linewidth]{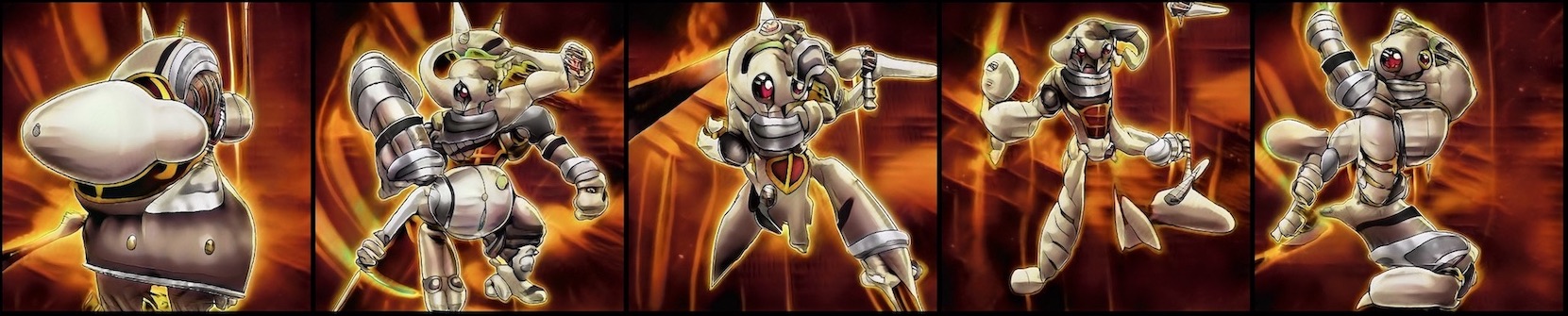}
  \caption{}
  \label{fig:probA}
\end{subfigure}%
\begin{subfigure}[b]{0.5\linewidth}
  \centering
  \includegraphics[width=0.9\linewidth]{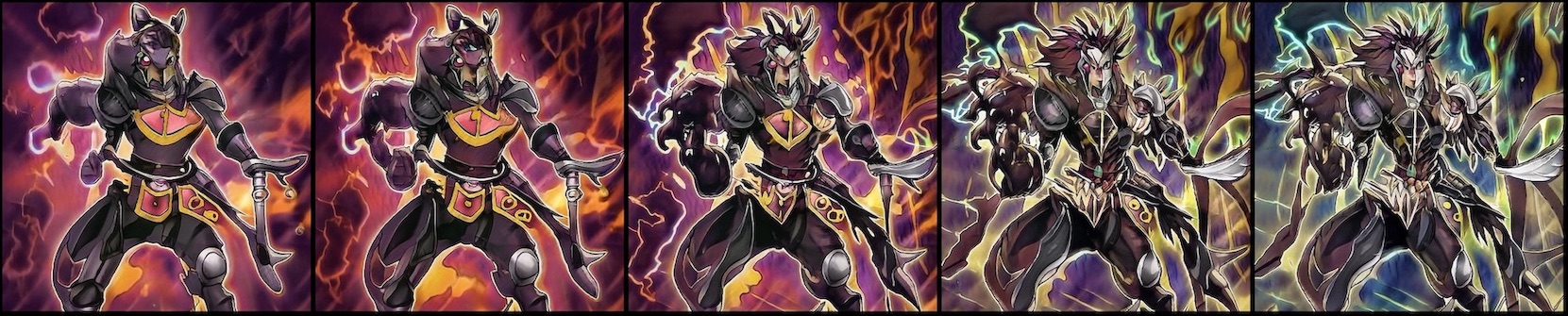}
  \caption{}
  \label{fig:probB}
\end{subfigure}
   \caption{StyleGAN2 does not always behave as desired in the low-data regime, and coarse-scale noise variables appear to be the problem. (a) Changing coarse-scale noise variables can cause large variations in identity, which is undesirable in practice. We show samples produced by a standard StyleGAN2. Samples are obtained by fixing latent variables, and using distinct noise instances. We expect images whose content is consistent, but instead the samples vary quite strongly. (b) Coarse-scale noise variables interfere with content control. We show samples obtained by varying a latent variable along a principal component of the latent distribution (as in~\cite{shen2020closed}). The expected behavior is a strong change in content, with minor style changes; but the samples vary substantially in style, and minimally in content.}
\label{fig:prob}
\end{figure*}


Until recently, hundreds of thousands of samples were required for GAN convergence. However, in the past year multiple labs independently converged on a set of data augmentations that drastically reduces the amount of data required (to as low as a few thousand or even a few hundred)~\cite{Karras2020ada,zhao2020differentiable}. These augmentations are called discriminator augmentations since they are applied to discriminator inputs instead of generator inputs (preventing augmentations from ``leaking'' into the generator). Augmentations like color jittering, affine transformations, cutout, and noise can be applied with fixed probability or adaptively. We found modest improvements in visual fidelity with adaptive discriminator augmentations (ADA) as compared to without. Our experiments also suggest that ADA is generally superior to augmenting with fixed probabilities on this dataset. We only use blitting, geometric, and color transforms as suggested in the ADA literature.



\section{Training the GAN \& Experimentation}
\label{sec:train}
For small, diverse datasets, we show that our proposed training procedure that suppresses noise in coarse-scale layers is superior to previous techniques in three ways fundamental to real-world use cases: (\ref{sec:betSyn}) higher quality synthesis; (\ref{sec:fine}) more stable exploration in the neighborhood of an existing image; and (\ref{sec:control}) stronger control over content.






\subsection{Better Synthesis}
\label{sec:betSyn}

We evaluate StyleGAN2 configurations via FID, summarized in Table~\ref{table:fid1}. All FID's are computed comparing 50k synthesized samples with the monsters-only training set of 6800 images (which the variants were trained on).  

We compare a standard StyleGAN2 with our model, which uses noise weights fixed to $0$ for coarse-scale noise during training and inference (specifically, layers of resolution $4^2-32^2$). Our model is significantly better for card-art. We believe our simple modification obtains superior synthesis quality because in the standard model, there are insufficient samples to control the coarse-scale noise weights. As a result, the coarse-scale noise becomes entangled with the latents
and both affect the long-scale structure of the image (see Fig.~\ref{fig:prob}). Our method forces the model to control long-scale image behavior using only latents and not noise.

\textbf{Our procedure produces better models for card art.}
A standard StyleGAN2 network (noise in all layers) produces a FID of $12.75$; our model produces a FID of $10.73$.

\textbf{Noise has important effects for card art.}
The importance of noise is documented in Fig. 5 of the original StyleGAN work~\cite{Karras_2019_CVPR}. Noise buffers were shown to enrich high frequency details at high-resolution layers, and can enable more complex low-frequency features at the low resolution layers~\cite{feng2021understanding}. Noise is important for card art, too: a StyleGAN trained with no noise in all layers produces a FID of $72.82$ (which
is very bad; Table~\ref{table:fid1}). Qualitatively, we found that the latent maintained complete control over the synthesis, but we observed flatter textures and overall reduction in richness and dynamic range that is characteristic of card art.

\textbf{For standard models, fixing noise at inference is harmful.}
A standard StyleGAN2 network (noise in all layers) produces a FID of $12.75$. This network relies on random noise at inference time. By modifying inference such that each latent variable uses the same noise buffers, we obtain a much worse FID of $34.42$. This suggests strongly that the noise is controlling long-scale aspects of the synthesis. 

\textbf{For our model, fixing noise at inference has minimal effect.}
Our model produces a FID of $10.73$ with constant per-latent noise, and $10.75$ with random per-latent noise - both of which significantly exceed the previous method.

\begin{table}[]
\begin{tabular}{||c | c ||}
 \hline\hline
 \textbf{Configuration} & \textbf{FID} \\
 \hline\hline
 No Noise  & $72.82$ \\
 \hline\hline
 Noise all layers, constant noise at inference  & $34.42$  \\
 \hline
 Noise all layers, different noise at inference  & $12.75$ \\
 \hline\hline
 Noise in layers above 32-res, const noise & $10.73$  \\
 \hline
 Noise in layers above 32-res, random noise & $10.75$ \\
 \hline\hline

\end{tabular}
\caption{FIDs for various network configurations when comparing 50k synthesized samples against the training set. No noise performs worst (first row). The previous off-the-shelf technique of adding noise to all layers is the next best (rows 2 and 3). In this case, changing the noise from constant for each sample (row 2) to different (row 3) dramatically improved FID, suggesting that selection of noise materially affects the generated distribution. Our proposed method of disabling noise in coarse-scale layers (specifically, 32-res and lower, rows 4 and 5) best captures the training data and offers superior GAN editing capabilities.}
\label{table:fid1}
\end{table}

\subsection{Fine-Scale Exploration}
\label{sec:fine}
\begin{figure}[h]
\begin{subfigure}[b]{\linewidth}
  \centering
  \includegraphics[width=0.9\linewidth]{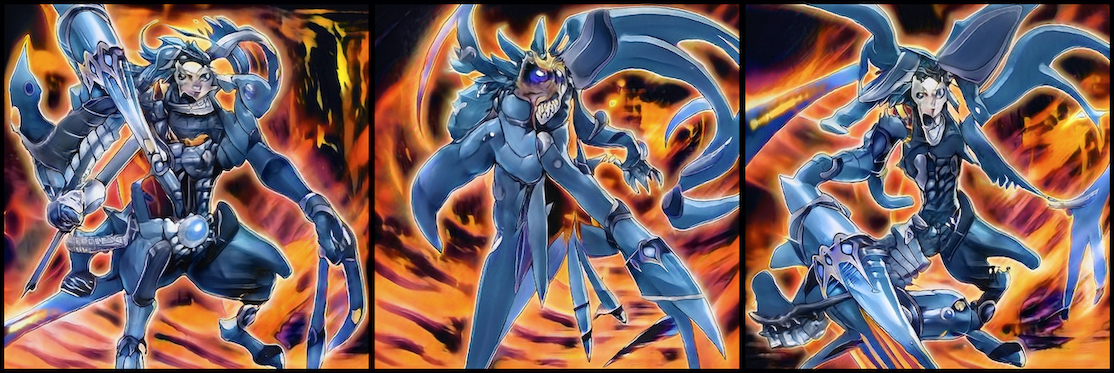}
  \caption{}
  \label{fig:noiseDiffOld}
\end{subfigure}%
\\
\begin{subfigure}[b]{\linewidth}
  \centering
  \includegraphics[width=0.9\linewidth]{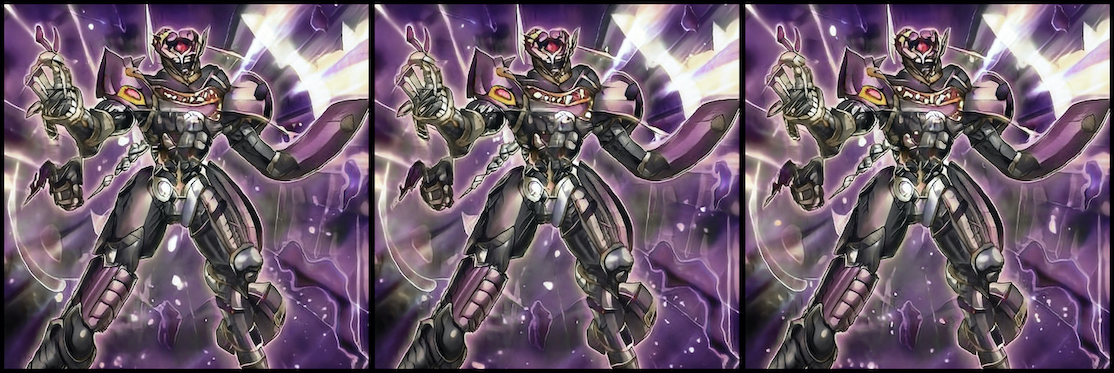}
  \caption{}
  \label{fig:noiseDiffNew}
\end{subfigure}
   \caption{Our approach of suppressing coarse-scale noise variables during training results in less aggressive local exploration, which is more useful in practice. (a) Samples produced by a standard StyleGAN2. Samples are obtained by fixing latent variables, and using distinct noise instances, resulting in large content changes and minor style changes. An artist cannot use noise changes to explore minor variations in card appearance because the changes are too drastic. (b) Samples with the same latent and different noise realizations produced by our modified GAN show small high frequency changes in appearance (e.g. white dots between the character's knees), allowing local exploration.}
   
\label{fig:noiseDiff}
\end{figure}

It is useful for artists to explore the data manifold near a sample, which changing the noise can achieve. We show different noise realizations for the same latent in Fig.~\ref{fig:noiseDiffOld} (previous network, noise all layers) and Fig.~\ref{fig:noiseDiffNew} (new network, no noise in coarse layers). Using the old technique, the identity of the creature dramatically changes based on the particular noise parameters. In contrast, creature identity is maintained with our approach. Only high frequency details change when updating the noise. 

Thus using an off-the-shelf network would give unpredictable results when attempting local exploration, which is undesirable for artists. Our proposed network does not suffer from this problem. Though we acknowledge that only changing finer-scale noise parameters could improve local exploration of the old network, synthesis quality and control are still superior with ours (Sections \ref{sec:betSyn} and \ref{sec:control}). 

To further validate that our proposed network is controlled by the latent and not noise, we generate 50k latents, and two sets of noise per latent. The first set of images uses constant noise for all latents, the second set of images uses random noise per latent. Thus, the only difference between the two sets of generated distributions is the selection of noise buffers, not latents. We perform this experiment independently for both the old and new networks. After comparing the two distributions generated for each network, the traditional network with noise in all layers obtained a FID of $21.1$, whereas the new network without noise in low res layers produced a FID of $0.221$, suggesting that our new method generates images far less sensitive to particular noise instances.





\subsection{Better Control} 
\label{sec:control}

\begin{figure}
\begin{subfigure}{\linewidth}
  \centering
  \includegraphics[width=\linewidth]{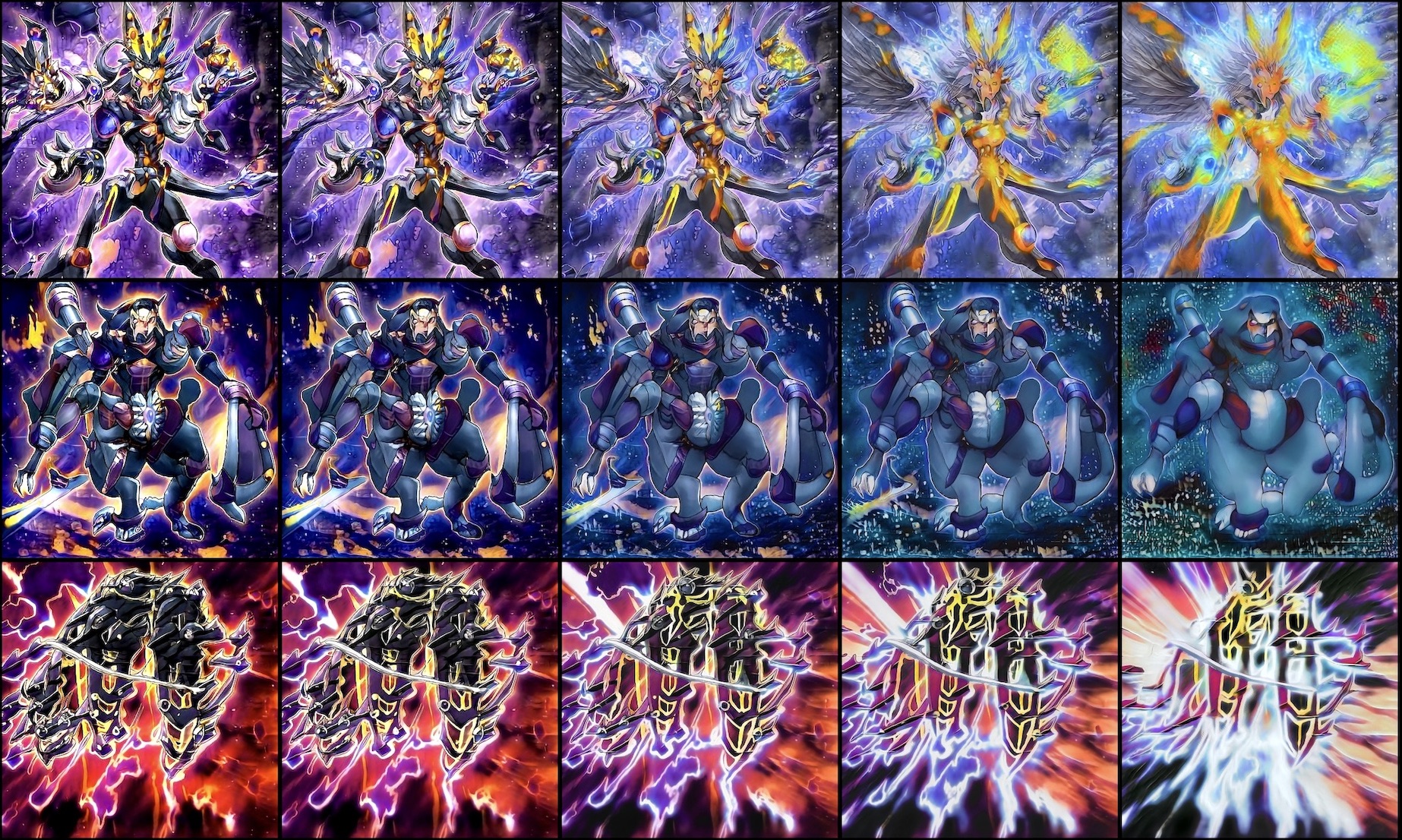}
  \caption{Latent variable control of identity is weak in StyleGAN2 in sparse, low-data settings. The center column shows three samples from StyleGAN2 (FID: 12.75) trained on our data. Columns to the left (resp. right) show the effects of changes of fixed size in the first principal component of the latent variable (as in~\cite{shen2020closed}); desired behavior is a change in identity. In practice, this component is changing style.}
  \label{fig:oldPCA} 
\end{subfigure}%
\\
\begin{subfigure}{\linewidth} 
  \centering
  \includegraphics[width=\linewidth]{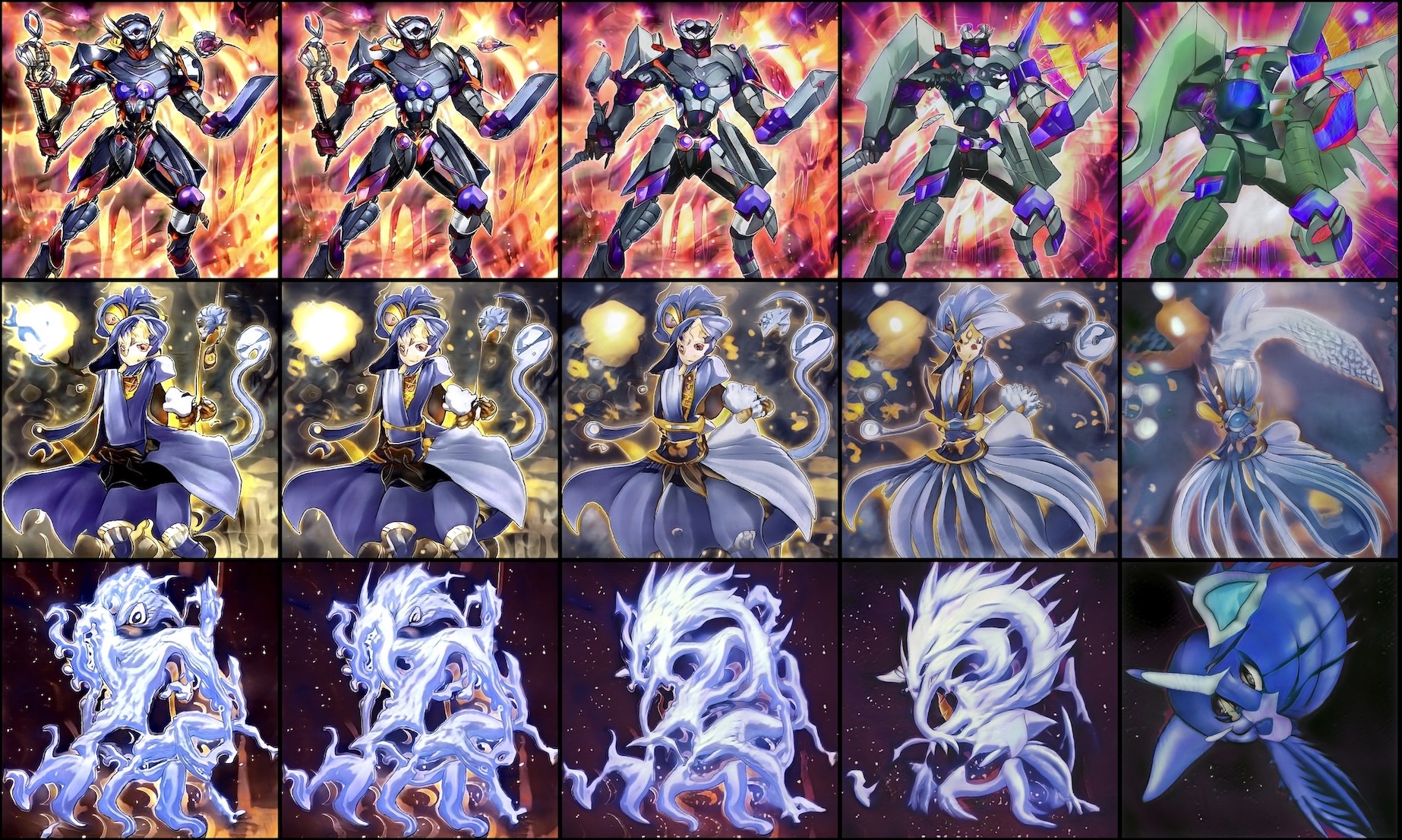}
  \caption{Suppressing coarse-scale noise improves latent variable control of identity. The center column shows three samples from our variant of StyleGAN2 trained on our data (FID: 10.7). Columns to the left (resp. right) show the effects of changes of fixed size in the first principal component of the latent variable. Note how the identity of the depicted creature is changing strongly.  PCA increments have the same magnitude as in (a).}
  \label{fig:newPCA}
\end{subfigure}
\\
\begin{subfigure}{\linewidth}
  \centering
  \includegraphics[width=\linewidth]{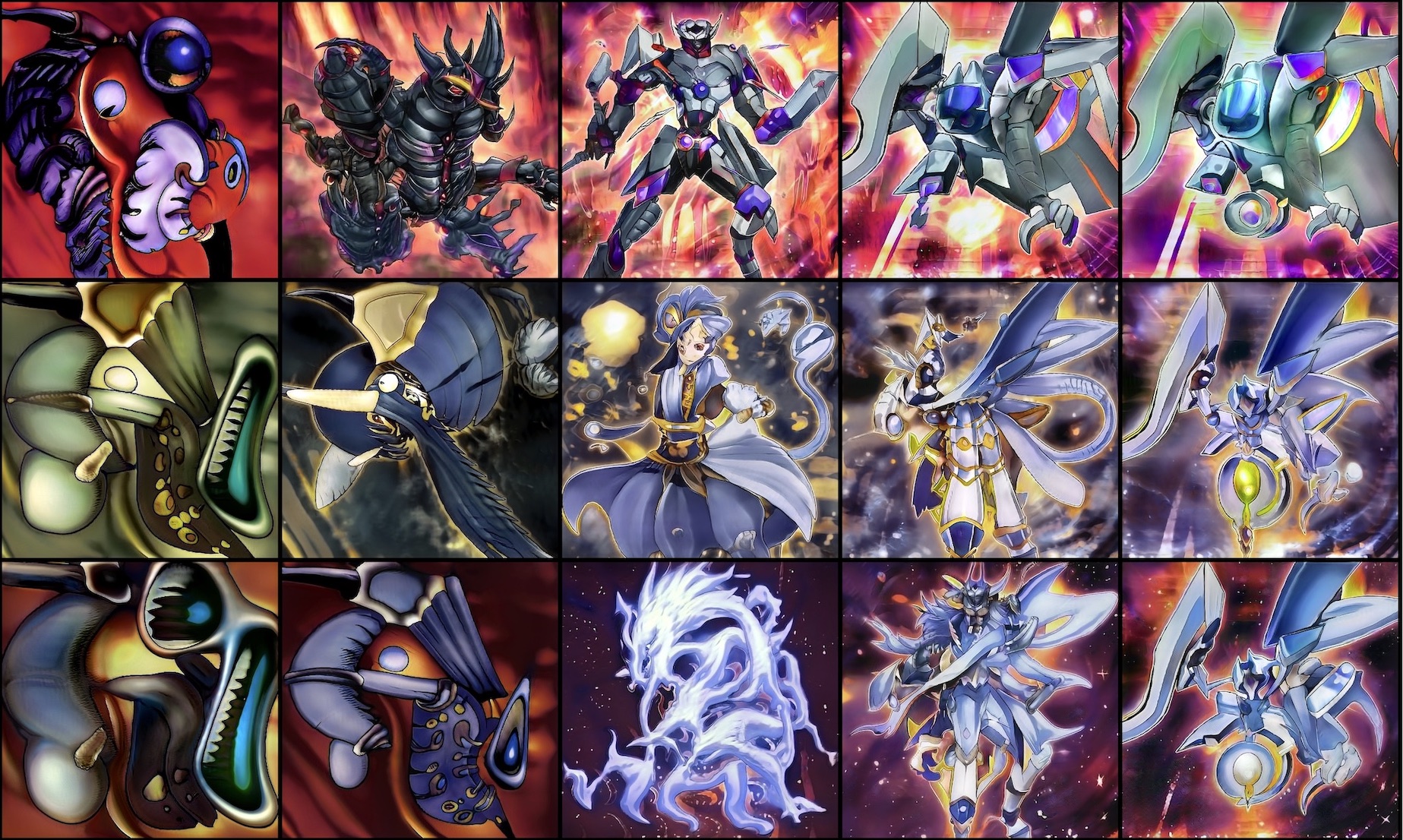}
  \caption{Same network as (b). Improvements in latent variable control apply to other principal components, producing a different set of changes that significantly change identity in contrast to (a)}
  \label{fig:newPCA2}
\end{subfigure}
\caption{Latent PCA results.}
\label{fig:pca}
\end{figure}




Recent work has analyzed the GAN latent space and network weights ~\cite{2020ganspace,shen2020closed}. By performing PCA, perturbing latent codes in the directions of eigenvectors was shown to correspond with meaningful changes in the resulting image. For FFHQ (human faces), gender, hair style, skin color and much more can be independently edited using PCA. In the context of sparse card art, PCA affects the synthesis in a complex, at times input-dependent way as we show in Fig.~\ref{fig:pca}. Simply using previous techniques drastically diminishes the power of PCA to influence the input as shown in Fig.~\ref{fig:oldPCA}, where PCA affects art style but not identity. This is undesired since we already have a mechanism to edit style while retaining identity (style-mixing, Section~\ref{sec:retain}). Instead, by training a network without noise in low res layers, PCA exposes meaningful controls over creature identity as shown in Figs.~\ref{fig:newPCA} and ~\ref{fig:newPCA2}. Thus, PCA and style-mixing can work in harmony to edit creature style and identity, which would be much harder using previous methods.


\section{Retaining Key GAN Properties}
\label{sec:retain}

\begin{figure}[t]
  \centering
  \includegraphics[width=1\linewidth]{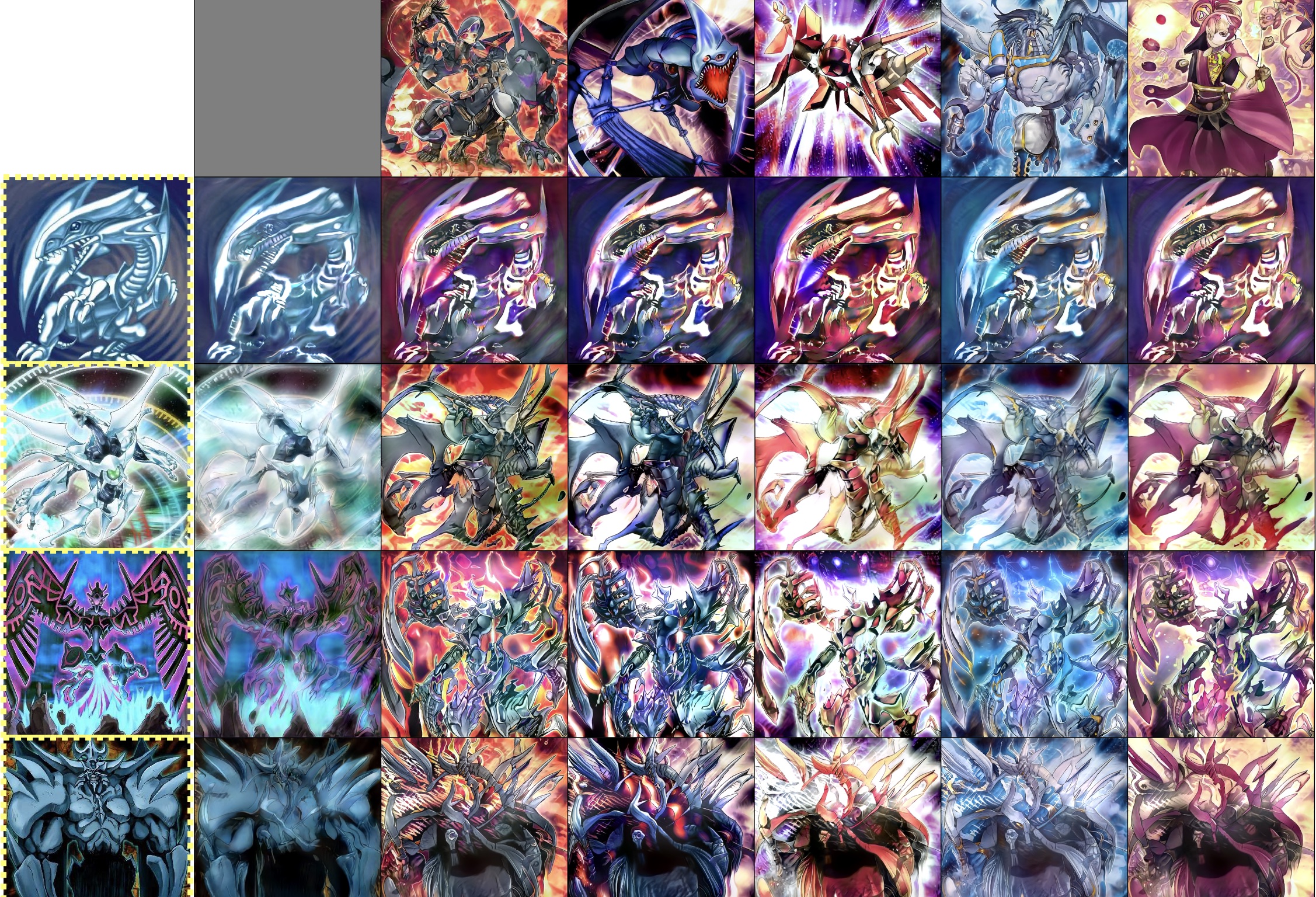}
  

   \caption{Suppressing coarse-scale noise does not affect other important editing functions. Here we show that projection and style-mixing still work in our approach. The first column in dotted lines shows four images from the training set. The second column shows the result of projecting those images into the latent space and recovering the corresponding latent code. The first row shows five synthesized images with known latents. The remaining images show the result of style-mixing the fine-scale latents from the first image in that column with the coarse-scale latents from the first image in that row.}
\label{fig:projMix}
\end{figure}


\textit{Latent Projection}: One key capability of StyleGAN is recovering a latent code given a target image ~\cite{xia2021gan}. Previous work has done so quite successfully via an optimization-based routine ~\cite{huh2020ganprojection}, favoring solutions that maintain fidelity after editing. We show projection of images from the Yu-Gi-Oh dataset and subsequent style-mixing in Fig.~\ref{fig:projMix}.



\textit{Style-Mixing}: Because StyleGAN builds images in layers of increasing resolution, with each layer controlled by a latent code, it is possible to mix the high res latents from one image, defining the style, and the low res latents from another image, defining the identity. We show that Yu-Gi-Oh art possesses this capability in Fig.~\ref{fig:projMix}. Thus, in an artist workflow, style-mixing enables creators to generate new art that is consistent in style.

\section{Artist Workflow}
\label{sec:aw}

We have thus far described a neural network capable of generating images in the manifold of card art. We now show how to deploy the network such that it can be used by artists in a production environment. 

For optimal interactive speeds, we recommend deploying on a machine with a GPU. Considerable advancements have been made in network quantization and pruning which can accelerate inference speeds on any hardware \cite{blalock2020state,zhou2017incremental} though we do not test them here. Running inference on an A100 GPU takes 0.05 seconds per image on average, suitable for interactive performance. 

For interactive viewing and editing of images, we recommend a GUI like streamlit, which can load, run, and display StyleGAN2 outputs with minimal code. Streamlit also makes it easy to create sliders for editing various network parameters as well as  downloading/uploading latent codes associated with images to ensure reproducibility. Some use cases may require deploying multiple models - a simple dropdown in streamlit enables easily selecting between multiple pretrained models.


The GUI can expose several controls for artists. A random seed changes the initial latent code. Another seed is associated with the noise buffers, which change high frequency details as shown in Fig.~\ref{fig:noiseDiffNew}. A slider could be exposed for latent truncation, a common technique to reduce sample diversity in exchange for sample fidelity, by moving the latent towards the mean of the latent distribution.

For style-mixing, we will need a scalar between 0 and the number of layers as the cutoff between low res latents from one image, and high res latents from another. We will need a random seed for the high res latents to mix with, and a percentage of style-mixing to apply between the source and target latent codes for mixing (only at the high res layers). 

To edit the latent codes in semantically meaningful directions, we can introduce a handful of sliders, each of which corresponds with a PCA direction for the latents. For a 512 dimensional latent code, there will be 512 PCA directions. To maintain a clean GUI, perhaps only 10  (or so) PCA sliders at a time can be presented to the user, with the user having the ability to edit the PCA index of any of the sliders. 

Finally, to ensure reproducibility, buttons should be exposed to download the latent and noise associated with the current image, and upload a latent code (these can be saved in .npz format). We list all these parameters in Table \ref{table:listParams}. 

\begin{table}[]
\begin{tabular}{||c | c | c ||}
 \hline
 \textbf{Param Name} & \textbf{dtype} & \textbf{range} \\
 \hline\hline
 Latent Seed  & int64 & $[0,max(int64)]$ \\
 Noise Seed  & int64 &  $[0,max(int64)]$ \\
 Truncation  & float32 &  $[-2,2]$ \\
 \hline
 Style-Mix Seed  & int64 & $[0,max(int64)]$ \\
 Style-Mix Cutoff & int32 & $[0,15]$ \\
 Style-Mix Strength & float32 & $[0,1]$ \\
 \hline
 PCA Direction (x10) & int32 & $[0,511]$ \\
 PCA Weight (x10) & float32 & $[-40,40]$ \\
 \hline
 Download Latent \& Noise & N/A & N/A \\
 Upload Latent \& Noise & N/A & N/A \\
 \hline
\end{tabular}
\caption{A list of parameters to control a deployed StyleGAN in a production environment.}
\label{table:listParams}
\end{table}

\section{Discussion \& Future Work}
\label{sec:disc}

Our overall assessment is that StyleGAN2 does an outstanding job capturing the vast array of styles - textures, lighting, and patterns -  of Yu-Gi-Oh cards, but shows some shortcomings in creating structurally coherent creatures with expressive high frequency details. The quality is not yet indistinguishable from the training set visually and quantitatively (we got a FID of 10.7). Constructing near-perfect deepfakes has essentially been done in other well-posed domains like human faces, anime, landscapes, and pets. These datasets are class-consistent, train on a higher volume of data, and produce denser data manifolds. 

Our finding that noise and latents are not sufficiently decorrelated using previous techniques, and subsequent workaround of training without noise in low resolution layers, is critical to obtaining editable results that could be used in a production workflow. We suspect the noise-latent trade-off issue does not manifest in the well-posed large-data regime, but will reappear in real-world low-data sparse contexts like our card art dataset. 

\begin{figure*}[t]
\begin{center}
\includegraphics[width=1\linewidth]{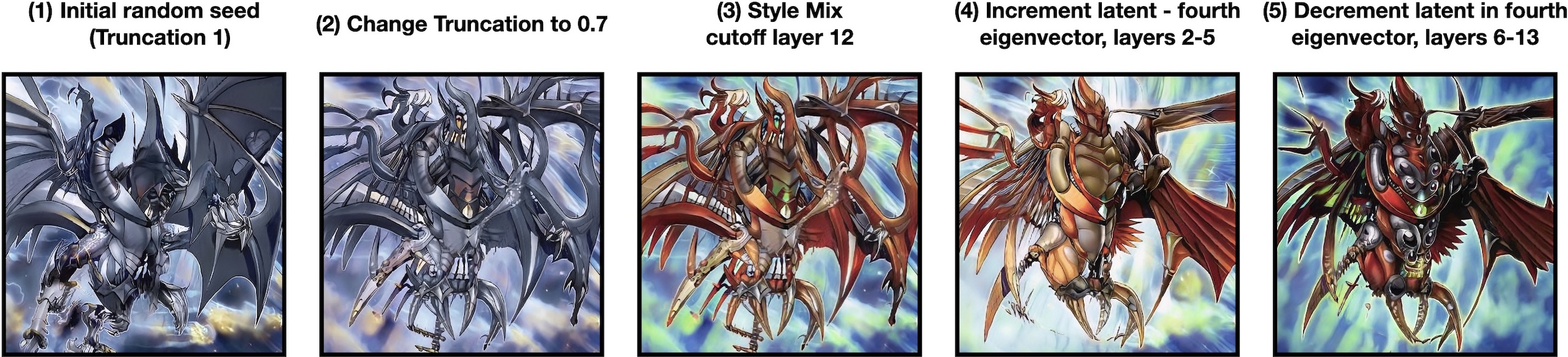}
\end{center}
   \caption{Sequentially editing a random StyleGAN2 output, resulting in a highly stylized creation. Edits (4) and (5) would have been much harder using previous techniques that fail to sufficiently decorrelate noise from latents as we address here.}
\label{fig:SeqEdit}
\end{figure*}


Recent work in generative modeling of sparse datasets has proposed leveraging a pre-trained StyleGAN trained on a similar domain to the (sparse) target dataset, and transferring the style ~\cite{li2020few,ojha2021few-shot-gan}. In the context of game art, such work shows promise, particularly in more complex artist workflows where multiple GAN models can be maintained for different data classes. 

This work does not condition the output based on any of the card attributes (aside from monsters-only), which follow-up work may consider.


One avenue for improvement could come from applying Instance Selection, which we hypothesize has failed due to the absence of a perceptually-aligned embedding. Attempting to train a classifier on card art (to predict some attribute of the card, for example) and using the classifier's features as an embedding function for Instance Selection could be a useful experiment to improve quality. 

We summarize our suggested art synthesis tool in Fig.~\ref{fig:SeqEdit}. We show how GAN paradigms like truncation, style-mixing, and latent PCA edits can be applied sequentially to create customized art. We emphasize that the final PCA edit shown in the figure would have been much harder using previous techniques that fail to sufficiently decorrelate noise from latents as we address in this work.


\section{Conclusion}
\label{sec:conc}
We have presented a new card art dataset challenging for generative networks to model. We showed that StyleGAN2 can produce compelling creature art with control, and analyzed the network's capabilities and limitations. In doing so, we demonstrated shortcoming of previous GAN methods, that noise and latents are overly entangled in challenging image domains, and we successfully addressed this problem here. We quantitatively and qualitatively showed that training on this card art dataset without noise in low res layers improves synthesis quality as well as subsequent editing capability. Finally, we proposed how the trained network can be deployed into an artist-friendly tool to assist in designing new creatures. 

{\small
\bibliographystyle{ieee_fullname}
\bibliography{egbib}
}

\end{document}